\documentclass{article}
\usepackage[letterpaper,margin=1in]{geometry}
\usepackage[round]{natbib}      
\usepackage{authblk}            

\makeatletter
\renewcommand\@maketitle{%
  \newpage \null
  \vskip -1.5em
  \begin{center}%
  \let\footnote\thanks
    {\LARGE \@title \par}%
    \vskip 1.2em%
    {\large \lineskip .5em \@author \par}%
    \vskip 0.8em%
    {\large \@date}%
  \end{center}%
  \par \vskip 1em}
\makeatother

\usepackage{amsmath,amssymb}
\usepackage{microtype}
\usepackage{graphicx}
\usepackage{booktabs}
\usepackage{url}
\usepackage{enumitem}
\setlist{nosep,leftmargin=*}
\setlist[itemize]{topsep=2pt,itemsep=1pt}
\setlist[enumerate]{topsep=2pt,itemsep=1pt}

\usepackage[hidelinks]{hyperref}
\usepackage{listings}
\let\oldthebibliography\thebibliography
\let\endoldthebibliography\endthebibliography
\renewenvironment{thebibliography}[1]{%
  \oldthebibliography{#1}%
  \setlength{\itemsep}{0pt plus 0.2pt}%
  \setlength{\parsep}{0pt plus 0.2pt}%
}{\endoldthebibliography}

\title{Loud or Silent? A Reusable Framework for Per-Modality\\
Failure Analysis in Multimodal Clinical AI}

\author[1,2]{Quang Bui}
\author[3]{Shlok Jaiswal}
\author[4]{Samuel Paik-Heintz}
\author[5]{Kevin Zhou}
\author[1,6]{Kaushik Madapati}
\author[7,8]{Krittaphas Chaisutyakorn}
\author[9,10]{Noah Dane Hebdon}
\author[11]{Dimitrios Proios}
\author[1]{Sebasti\'{a}n Andr\'{e}s Cajas Ord\'{o}\~{n}ez}
\author[1,19]{Kacper Dobek}
\author[1,11,12]{Boya Zhang}
\author[20]{Aly Dhedhi}
\author[1,12,13]{Ahram Han}
\author[14]{Kushul Reddy Palakala}
\author[1,15,16,17]{Rahul Gorijavolu}
\author[21]{Jacques Kpodonu}
\author[1,8,18]{Leo Anthony Celi}

\affil[1]{MIT Critical Data, Massachusetts Institute of Technology, Cambridge, Massachusetts, United States}
\affil[2]{American International School Vienna, Vienna, Austria}
\affil[3]{Neuqua Valley High School, Naperville, Illinois, United States}
\affil[4]{North Hollywood High School, Los Angeles, California, United States}
\affil[5]{Hopewell Valley Central High School, Pennington, New Jersey, United States}
\affil[6]{Department of Electrical Engineering and Computer Sciences, University of California, Berkeley, Berkeley, California, United States}
\affil[7]{Siriraj Informatics and Data Innovation Center, Faculty of Medicine, Siriraj Hospital, Bangkok, Thailand}
\affil[8]{Harvard T.H. Chan School of Public Health, Harvard University, Boston, Massachusetts, United States}
\affil[9]{Quantum Innovation Centre (Q.InC), Agency for Science, Technology and Research, Singapore}
\affil[10]{School of Advanced International Studies, Johns Hopkins University, Washington, District of Columbia, United States}
\affil[11]{Department of Radiology and Medical Informatics, University of Geneva, Geneva, Switzerland}
\affil[12]{Institute for Medical Engineering and Science, Massachusetts Institute of Technology, Cambridge, Massachusetts, United States}
\affil[13]{Department of Surgery, Seoul National University Hospital, Seoul, South Korea}
\affil[14]{University of North Florida, Jacksonville, Florida, United States}
\affil[15]{School of Medicine, Johns Hopkins University, Baltimore, Maryland, United States}
\affil[16]{Department of Biomedical Engineering, Johns Hopkins University, Baltimore, Maryland, United States}
\affil[17]{Artificial Intelligence for Responsible, Generalizable, and Open Surgical (ARGOS) Research Group, Baltimore, Maryland, United States}
\affil[18]{Division of Pulmonary, Critical Care and Sleep Medicine, Beth Israel Deaconess Medical Center, Boston, Massachusetts, United States}
\affil[19]{Institute of Computing Science, Poznan University of Technology, Poznan, Poland}
\affil[20]{American Heritage School, Plantation, Florida, United States}
\affil[21]{Division of Cardiac Surgery, Beth Israel Deaconess Medical Center, Harvard Medical School, Boston, Massachusetts, United States}
\date{}

\begin{document}
\maketitle
\vspace{-2.5em}
\begin{abstract}
Multimodal clinical models are usually judged on accuracy with every modality present, but
deployment removes modalities; an echocardiogram is often unavailable where an ECG is routine.
Two questions then matter beyond the size of the accuracy loss: which modality was responsible,
and whether the model fails \emph{loudly} or \emph{silently} once that modality is dropped. The
distinction is per-example and modality-level, and is separate from post-hoc feature attribution
(e.g.\ SHAP). Models are replaced often; the evaluation that answers these questions is reused.
We present a model-agnostic \emph{modality-failure framework}: given $N$ modality embeddings,
any mask-aware probe, and labels, it returns a per-example failure taxonomy, a per-modality
complementarity matrix that attributes error to modalities, and a loud-vs-silent dropout profile
separating monitorable failures from those that pass unflagged far from the decision boundary,
using only deployment-observable signals. We release it as a small, unit-tested harness and
validate it against planted ground truth. Across seeds it recovers that planted
modality dominance and complementary subset, reports per-modality loud-vs-silent rates, and
scales to a three-modality complementarity matrix; because the planted structure is known by
construction, this validates recovery of per-example attribution rather than clinical
performance. We then instantiate the framework on \emph{frozen} EchoJEPA and HuBERT-ECG
embeddings for LVEF and the \mbox{EF\,$\leq$\,40\%} HFrEF gate over a paired MIMIC-IV cohort,
where on the held-out test split (\mbox{$n=245$}) dropping echo nearly doubles error. The paired cohort and both frozen
embedding sets are already in hand, so the per-example attribution and loud-vs-silent profile on
real data are a direct next run rather than an access-limited one. The narrow
echo-to-ECG overlap that bounds cohort size is itself a deployment finding for
cardiac foundation models. All of our work can be found at \url{https://github.com/criticaldata/PRIMED-AI}.
\end{abstract}

\newpage

\section{Introduction}
\label{sec:intro}
Foundation models for medical imaging and signals are replaced every few months, while the
evaluation that exposes \emph{how} they fail is reused across model generations. Multimodal
clinical models are still judged mostly on accuracy with every modality present. Deployment
rarely obliges. A model fusing echo and ECG may be trained on both, but at inference the ECG is
acquired almost everywhere, while the echo needs equipment and expertise concentrated in
well-resourced settings. When a modality goes missing, three questions matter that an accuracy
number cannot answer. Which modality was responsible for the error? Does the model fail
\emph{loudly}, its output signalling doubt where a monitor can catch it, or \emph{silently}, far
from the decision boundary yet wrong and so unflagged? And how do failures distribute across
patients?

We therefore argue for a \emph{failure-analysis framework} over yet another fusion model, and
provide one (Table~\ref{tab:pillars}). It is model-agnostic: given $N$ modality embeddings, any
mask-aware probe, and labels, it returns a per-example failure taxonomy, a per-modality
complementarity matrix, and a loud-vs-silent dropout profile, so it transfers across model
generations. We instantiate it on \emph{frozen} EchoJEPA and HuBERT-ECG embeddings (no fine-tuning, a
realistic stand-in for how a small team builds such a system) for LVEF regression and the
\mbox{EF\,$\leq$\,40\%} HFrEF gate over MIMIC-IV, validate that it recovers planted structure on
data with planted ground truth, and demonstrate it on a real paired cohort. Scope is deliberately narrow
(LVEF / \mbox{EF\,$\leq$\,40\%} only; valvular and HFpEF deferred for label reasons).

\paragraph{Contributions.}
\begin{itemize}
  \item A model-agnostic modality-failure framework, with formal definitions for a per-example
        failure taxonomy, per-modality complementarity and attribution, and a loud-vs-silent
        dropout profile that uses only deployment-observable signals.
  \item A small reusable harness implementing it for any $N$ modalities and any mask-aware probe,
        validated against planted ground truth. Across 12 seeds it recovers the planted dominance
        and complementary subset, and it scales to an $N$-modality complementarity matrix.
  \item An instantiation on frozen echo and ECG embeddings for LVEF/HFrEF over MIMIC-IV, with a
        real-data degradation demonstration on the full paired cohort, together with the
        observation that the modality expensive to obtain at inference is also the expensive one to
        build infrastructure for, which we read as a deployment finding for cardiac foundation
        models.
\end{itemize}

\begin{table}[!htbp]
\centering
\caption{What the framework reports, and the deployment question each view answers.}
\label{tab:pillars}
\small
\begin{tabular}{@{}p{0.26\linewidth}p{0.64\linewidth}@{}}
\toprule
\textbf{View} & \textbf{Question it answers} \\
\midrule
Failure taxonomy & Are errors merely imprecise, or \emph{critical} (the clinical gate flips)? \\
Complementarity matrix & \emph{Which} modality carries the signal, and where are modalities redundant vs.\ complementary? \\
Loud-vs-silent profile & When a modality is missing, does the wrong prediction land near the decision boundary (loud, monitorable) or far from it (silent, unflagged)? \\
\bottomrule
\end{tabular}
\end{table}

\section{Related Work and Positioning}
\label{sec:related}
Foundation models exist for both modalities: EchoJEPA~\citep{echojepa} adapts the V-JEPA\,2
latent-predictive video approach~\citep{vjepa2} to echocardiography, and HuBERT-ECG~\citep{hubertecg}
is a self-supervised model for the 12-lead ECG (HuBERT \citep{hubert2021} / wav2vec 2.0 \citep{wav2vec2} family). EchoingECG~\citep{echoingecg} distills echocardiographic knowledge into ECG
representations (probabilistic cross-modal embeddings with an echo--language teacher) so cardiac
function can be predicted from the ECG alone, with uncertainty, in zero-/few-shot and fine-tuned
settings. Our study differs on three deployment-relevant axes: (i)~both backbones stay
\emph{frozen} and we learn only lightweight probes, without distillation or fine-tuning; (ii)~we
\emph{retain and fuse} both modalities and ask what happens when one is \emph{missing at
inference}, rather than compressing echo into an ECG-only model; and (iii)~inference-time degradation is a primary result, and we prespecify a demographic fairness audit
to apply to full per-example predictions. EchoingECG asks how to build a
better ECG model using echo at training time; we ask, model-agnostically, how a deployed
multimodal model fails when a modality is gone. Missing-modality robustness is typically reported
as a single aggregate accuracy drop~\citep{smil2021,mmrobust2022}; we instead attribute failures per modality and separate loud
from silent degradation, packaged as a reusable instrument rather than a one-off ablation. This is
distinct from post-hoc feature-attribution methods (e.g.\ SHAP): those explain which input features
drove a single model's prediction, whereas loud-vs-silent operates under actual modality masking
and asks where the resulting per-example error lands relative to the clinical decision boundary.

\section{A Framework for Modality-Failure Analysis}
\label{sec:framework}
We formalize a model-agnostic analysis that assumes only a \emph{mask-aware} predictor: given
modalities $\mathcal{M}$, a held-out set with regression target $y$ (LVEF) and clinical gate
$g=\mathbb{1}[y\le\tau]$ ($\tau=40$, the HFrEF cut point \citep{heidenreich2022}), the HFrEF referral decision (correct when the model's gate
call matches the patient's true EF\,$\leq$\,40\% status, and \emph{flipped} when it does not), and a
function $f(S)$ returning a prediction per example
when only the subset $S\subseteq\mathcal{M}$ is present (others masked at inference), it reports
three views. Nothing else about $f$ is assumed, so it applies to any probe, fusion architecture, or
backbone.

\paragraph{(i) Failure taxonomy.} Each example, under each condition, is labelled \emph{correct}
($g$ right, $|f-y|\le\delta$), \emph{imprecise} ($g$ right, $|f-y|>\delta$), or \emph{critical}
($g$ wrong: an HFrEF case missed or false-alarmed). Critical errors are the clinically actionable
failures, and raw mean absolute error (MAE) does not isolate them; a large regression error that
does not flip the gate stays \emph{imprecise} by design.

\paragraph{(ii) Complementarity and attribution.} Each modality's leave-one-out value is
$\Delta_m{=}\mathrm{MAE}(f(\mathcal{M}{\setminus}\{m\}))-\mathrm{MAE}(f(\mathcal{M}))$, and per
example we attribute the ``win'' to the modality whose removal hurts that example most. Solo and
pairwise errors form a \emph{complementarity matrix} separating \emph{redundant} modalities (either
alone suffices) from \emph{complementary} ones (both needed). Views stratify by patient attributes
(sex, age, race): a modality that wins overall may not win within a subgroup, tying attribution to
equity.

\paragraph{(iii) Loud-vs-silent dropout profile.} When $m$ is missing, an example that was
gate-correct with $\mathcal{M}$ but is now gate-wrong is an \emph{induced critical failure}. We split
these by a deployment-observable signal, the prediction's margin from the decision boundary: a
failure is \emph{silent} if the wrong prediction sits far on the wrong side ($|f-\tau|\ge\kappa$, so
it looks unremarkable and a margin monitor would not flag it) and \emph{loud} if it lands near the
threshold ($|f-\tau|<\kappa$, where the output invites a second read). Margin is a \emph{proxy for
monitorability} rather than calibrated confidence (a probabilistic head, cf.\ EchoingECG, would
sharpen it), and it uses only the deployed prediction, never the unavailable full-modality
counterfactual. A high silent rate is the dangerous regime: wrong, far from the cut, and unflagged.

All three views are computed from a single call to the released harness.

\section{Data and Cohort Design}
\label{sec:data}
Having defined the modality-failure framework in model-agnostic terms, we now instantiate it in a cardiac setting where missing modalities are clinically realistic: frozen echocardiogram and ECG embeddings for LVEF estimation~\citep{lang2015} and the EF\,$\leq$\,40\% HFrEF gate.

We draw on three PhysioNet collections~\citep{goldberger2000,physionet2026}: MIMIC-IV-Echo, for
DICOM studies from v0.1~\citep{mimicivecho} and structured LVEF measurements from
v1.0~\citep{mimicivecho_v1}; MIMIC-IV-ECG, for 12-lead waveforms~\citep{mimicivecg}; and MIMIC-IV
v3.1 for demographics~\citep{mimiciv,johnson2023mimiciv}. Cohort construction runs server-side on
BigQuery~\citep{mimiccloud}, so the full corpus of roughly 525K DICOM videos across 7,243 studies
from 4,579 patients is never materialized locally.

\paragraph{LVEF labels.} LVEF is not stored in a single column, since lab-system transitions and
rest/stress protocols yield several structured measurement identifiers~\citep{mimicivecho_v1}. We
prefer a direct resting measurement where one exists, falling back to the biplane estimate, then to
the explicitly rest-labeled fields, and finally to the 3D estimate; ranges are used only when no
point estimate is recorded. Stress measurements are excluded, since they describe a different
physiological state. For auditability we store which field each value came from. From this we derive
continuous LVEF and the \mbox{EF\,$\leq$\,40\%} label.

\paragraph{Pairing and splits.} For each labeled echo study we match the nearest ECG from the same
patient within a symmetric $\pm$24\,h window, resolving multi-match cases to the nearest timestamp
with deterministic tie-breaking. Two alternatives are configurable and matter for how a deployment
is framed: an asymmetric window, admitting an ECG up to 30 days \emph{before} the echo but only
24\,h after, mirrors a screen-first setting in which the ECG is the earlier and cheaper test; and
encounter-aligned \emph{admission} pairing is robust to documented ECG clock skew. Splits partition
by patient (70/10/20) with an explicit zero-leakage check, and demographics (sex, anchor age, race,
and derived age bands) join for the fairness audit, with coverage and MIMIC gender-curation
limitations logged. Table~\ref{tab:cohort_funnel} reports the construction funnel. Temporal pairing
is the binding constraint by a wide margin: 6,324 echo studies have a subject-level ECG match, but
only 1,678 have one within 24\,h, and requiring an enclosing hospital admission for demographics
leaves 1,208 studies from 1,003 patients. A framework about missing modalities therefore meets a
cohort that is itself defined by how often the two modalities fail to co-occur.

\begin{table}[t]
\centering
\caption{Paired echo and ECG LVEF cohort construction funnel.}
\label{tab:cohort_funnel}
\small
\begin{tabular}{@{}cp{0.5\linewidth}rr@{}}
\toprule
\textbf{Step} & \textbf{Inclusion criterion} & \textbf{Studies} & \textbf{Patients} \\
\midrule
1 & Echo studies (all)                                  & 7{,}243 & 4{,}579 \\
2 & Linked to a structured measurement                  & 7{,}182 & 4{,}552 \\
3 & Patient has an LVEF reading                         & 6{,}916 & 4{,}472 \\
4 & Patient also has an ECG (subject overlap)           & 6{,}324 & 4{,}002 \\
5 & ECG within $\pm$24\,h window of the echo            & 1{,}678 & 1{,}326 \\
6 & Within a hospital admission (demographics attached) & 1{,}208 & 1{,}003 \\
\bottomrule
\end{tabular}
\end{table}


\section{Frozen-Embedding Pipeline}
\label{sec:method}


\subsection{Frozen encoders and probes}
This instantiation uses frozen EchoJEPA-L echo-video embeddings (ViT-L,
1024-dim)~\citep{echojepa}, released for MIMIC-IV-Echo on Hugging
Face,\footnote{\url{https://huggingface.co/datasets/MITCriticalData/mimic-iv-echo-jepa-embeddings}}
together with frozen HuBERT-ECG embeddings for the 12-lead ECG~\citep{hubertecg}. Both encoders run
in inference mode only, and each embedding is tagged with the checkpoint that produced it so results
stay traceable to an encoder version. Keeping the backbones frozen is what makes the framework
portable: only lightweight probes are trained, so the same evaluation applies unchanged to the next
released encoder.

Table~\ref{tab:probes} summarizes the four heads. Echo-only uses
\emph{attentive} pooling (a linear probe on raw high-dimensional EchoJEPA tokens is expected
to underperform). Cross-attention fusion is the primary model for missing-modality analysis:
bidirectional multi-head attention between echo and ECG token sequences, followed by an MLP
regression head. All probes predict continuous LVEF; the \mbox{EF\,$\leq$\,40\%} gate is derived
from predictions. Metrics: MAE and EF\,$\leq$\,40\% AUROC (Area Under the Receiver
Operating Characteristic) (context only, from cohorts not comparable to our split: EchoJEPA-L reports $5.97$ LVEF MAE on an internal Toronto cohort under frozen-backbone multi-view attentive probing \citep{echojepa}; an AI-ECG screen for EF $\le$ 35\% reports $0.93$ AUROC on a Mayo cohort \citep{attia2019}).

\begin{table}[t]
\centering
\caption{Probe architectures and roles.}
\label{tab:probes}
\small
\begin{tabular}{@{}lp{0.30\linewidth}p{0.34\linewidth}@{}}
\toprule
\textbf{Probe} & \textbf{Pooling / fusion} & \textbf{Role} \\
\midrule
ECG-only & mean or attentive & Ubiquitous-modality baseline; screen-first lower bound. \\
Echo-only & attentive & Upper bound when echo is available; attentive vs.\ linear ablation. \\
Concat-MLP & concatenated pooled branches into an MLP (each branch pooled as in its unimodal row above) & Lightweight fused baseline. \\
Cross-attn & bidirectional cross-attention & Primary fused model; supports branch masking. \\
\bottomrule
\end{tabular}
\end{table}

\subsection{Evaluation protocol}
The cross-attention probe is trained once on complete data, and the \emph{same deployed checkpoint}
is then evaluated on the held-out test split under three conditions: \emph{full}, \emph{echo-dropped}
(an ECG-only deployment), and \emph{ECG-dropped} (echo-only). Crucially, a branch is zeroed at
inference rather than retraining a unimodal model for each condition. Retraining would answer a
different question, namely how good a model built for one modality can be, whereas the deployment
question is what the model you already shipped does when an input goes missing. We report MAE and
\mbox{EF\,$\leq$\,40\%} AUROC per condition, with bootstrap confidence intervals where per-example
predictions are available, and plot the degradation curve. This is what separates graceful from
silent failure when the expensive modality is absent.

The fairness audit is post-hoc on those same test predictions and requires no additional training:
we stratify by sex, age band, and race, report per-stratum MAE and \mbox{EF\,$\leq$\,40\%} AUROC, and
flag strata falling below a minimum count rather than reporting an unreliable number for them.
Prespecifying the audit this way keeps it from becoming a search over subgroups. Throughout, runs are
configuration-driven with fixed random seeds, and each artifact carries a manifest recording the git
commit and resolved configuration that produced it.

\section{Results}
\label{sec:results}

\subsection{Validation: the harness recovers planted structure}
\label{sec:validation}
\begin{figure}[t]
\centering
\includegraphics[width=0.44\linewidth]{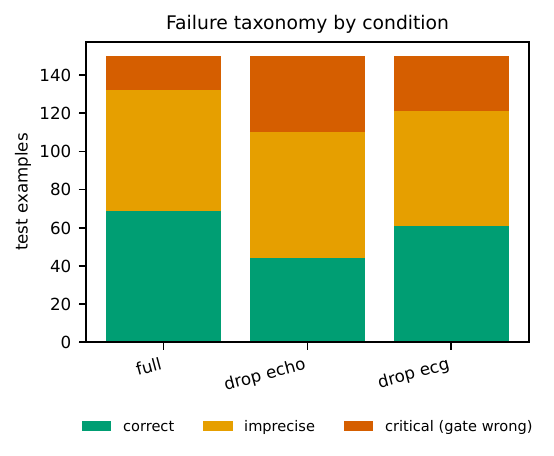}
\caption{Failure taxonomy by condition (representative seed; gate prevalence held healthy by
stratified splitting). Full-modality inference is mostly \emph{correct}; dropping echo introduces
many \emph{critical} gate-flipping failures. Counts, not just MAE, make the clinically actionable
errors explicit.}
\label{fig:taxonomy}
\end{figure}

We validate on synthetic echo/ECG data with planted structure, where echo encodes LVEF strongly and
ECG weakly, except in an $\approx$18\% subset where echo is uninformative so ECG must carry the
signal. We train one Ridge probe on the full embeddings and mask at inference; splits are
stratified by the gate and independent of data generation (test-set HFrEF prevalence held in a sane
band). This validation is circular by construction: because we plant the modality structure
ourselves, checking that the harness recovers it confirms the harness's mechanics, not that either
modality is clinically superior in real patients; this is nonetheless the only place a
per-example attribution ground truth exists, since no such label is available in real data. Across \textbf{12 seeds} the harness robustly recovers the planted
attribution: echo's leave-one-out value is $5.5{\pm}0.7$ MAE vs.\ $1.8{\pm}0.5$ for ECG, echo is the
more valuable modality in \textbf{all 12} seeds, and it takes $0.67{\pm}0.04$ of per-example wins
while ECG correctly wins on the planted complementary subset. The failure taxonomy
(Fig.~\ref{fig:taxonomy}) shifts from mostly \emph{correct} under full inference to many
\emph{critical} (gate-flipping) errors when echo is dropped. We also report the loud-vs-silent split
per modality; here the two are comparable ($0.31{\pm}0.07$ silent when echo is dropped vs.\
$0.32{\pm}0.10$ for ECG, with the ordering flipping across seeds). That is the intended behaviour:
value and detectability are \emph{separate} axes the harness measures, not an asymmetry it
assumes: the most valuable modality need not be the most dangerous to lose.

\subsection{Generality to \texorpdfstring{$N$}{N} modalities}
\label{sec:generality}
\begin{figure}[t]
\centering
\includegraphics[width=0.40\linewidth]{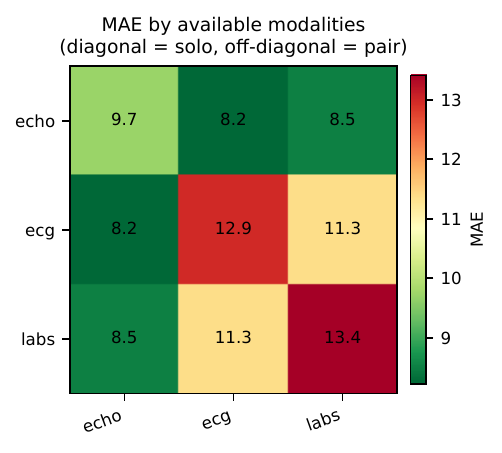}
\caption{Complementarity matrix on a three-modality instance: test MAE with each modality alone
(diagonal) and each pair (off-diagonal); lower is better. Every echo-containing combination far
outperforms ECG with labs, identifying echo as irreplaceable. Non-trivial attribution of this kind
requires \mbox{$N{\ge}3$}.}
\label{fig:matrix3}
\end{figure}

With only two modalities, per-example attribution is partly degenerate (dropping one equals using
the other), so we exercise the general path on a three-modality instance (echo, ECG, and a simulated
``labs'' channel; \mbox{$n{=}225$} test). The harness returns a genuine $3{\times}3$ complementarity
matrix (Fig.~\ref{fig:matrix3}). On the diagonal (each modality alone), echo achieves the lowest MAE $9.7$, vs.\ $12.9$ for ECG and $13.4$ for labs; off the diagonal, every echo-containing pair ($\approx$8.3) beats the ECG and labs pair ($11.3$). The marginal
values (echo $3.8$, ECG $0.9$, labs $0.7$) and per-example wins (echo $125$, ECG $56$, labs $44$) recover the planted ordering, flagging
echo as \emph{irreplaceable} and ECG and labs as weak and partly redundant. The same call scales to any $N$.

\subsection{Real cohort: an integration test}
\label{sec:realdemo}
Instantiated on real frozen EchoJEPA and HuBERT-ECG embeddings, a cross-attention probe on the
held-out test split (\mbox{$n{=}245$}, the test fold of the full paired cohort) reproduces the
expected aggregate dropout (Table~\ref{tab:degradation}): dropping echo (the ECG-only deployment)
nearly doubles MAE (from $10.28$ to $18.57$) in this single run. We treat this as an
\emph{integration test} of the end-to-end pipeline rather than a validation of the framework's
per-example views, since the taxonomy, complementarity, and loud-vs-silent outputs all require a
per-example prediction dump on real data. Producing that dump and running the released harness on it
is compute-cheap at this cohort size, and a canonical rerun with patient-level cross-validation is in
progress.

\begin{table}[t]
\centering
\caption{Real-cohort missing-modality degradation (single held-out run, \mbox{$n=245$}); same fused
checkpoint, one branch masked per condition.}
\label{tab:degradation}
\small
\begin{tabular}{@{}lcc@{}}
\toprule
\textbf{Inference condition} & \textbf{LVEF MAE}\,$\downarrow$ & \textbf{EF\,$\leq$\,40\% AUROC}\,$\uparrow$ \\
\midrule
Full (echo and ECG)        & \textbf{10.28} & \textbf{0.766} \\
Echo dropped (ECG-only)    & 18.57          & 0.693 \\
ECG dropped (echo-only)    & 15.13          & 0.750 \\
\bottomrule
\end{tabular}
\end{table}

\section{The Released Harness}
\label{sec:status}
The failure-analysis framework and its harness are the primary deliverable, and
Table~\ref{tab:release} inventories what is released. Every listed module carries unit tests, 51
passing at the time of writing, including tests that the harness recovers planted structure.

\begin{table}[!htbp]
\centering
\caption{Released components (representative).}
\label{tab:release}
\footnotesize
\begin{tabular}{@{}p{0.30\linewidth}p{0.62\linewidth}@{}}
\toprule
\textbf{Stage} & \textbf{Released artifact} \\
\midrule
Cohort &
BigQuery builder: pairing, LVEF hierarchy, demographics, funnel export, flowchart. \\
Splits &
Subject-level 70/10/20 manifest with zero-leakage assertion. \\
Prevalence &
\mbox{EF\,$\leq$\,40\%} prevalence overall and per split; AUROC reportability flag. \\
Encoders &
Frozen EchoJEPA-L and HuBERT-ECG encoder wrappers; HuBERT-ECG smoke-test script. \\
Extraction &
Batch echo/ECG forward passes into the embedding cache; cluster extraction scripts bundled. \\
Probes &
ECG-only, echo-only (attentive), concat-MLP, cross-attention (with branch masks). \\
Evaluation &
Missing-modality runner (bootstrap CIs); degradation-curve plotter; fairness stratifier. \\
Failure harness &
Model-agnostic \texttt{analyze(embeddings, predict\_fn, labels)}: failure taxonomy, complementarity matrix, loud-vs-silent dropout profile; figures + CLI. \\
\bottomrule
\end{tabular}
\end{table}

\paragraph{Reusable harness.} The framework is exposed as a small model-agnostic library: a single
\texttt{analyze} entry point takes $N$ modality embeddings, any mask-aware prediction function, and
labels, and returns the failure taxonomy, the complementarity matrix, and the loud-vs-silent dropout
profile. The same evaluator is therefore reused across probes, fusion architectures, and future
backbone releases; Listing~\ref{lst:api} shows the call.
\begin{lstlisting}[float=!htbp,language=Python,caption={Using the harness: any $N$ modalities, any mask-aware probe.},label=lst:api]
from primed_ai.failure import analyze_modality_failure

report = analyze_modality_failure(
    embeddings={'echo': echo_emb, 'ecg': ecg_emb},  # any N modalities
    lvef=y, ef_le_40=gate,                           # regression target + clinical gate
    predict_fn=probe.masked_predict,                 # subset -> predictions (others masked)
    groups={'sex': sex, 'age_band': age},            # optional: per-stratum attribution
)
report.complementarity['marginal_value']    # e.g. {'echo': 5.5, 'ecg': 1.8}
report.dropout['drop_echo']['silent_rate']  # loud vs. silent when echo is missing
report.conditions['drop_echo']['taxonomy']  # correct / imprecise / critical counts
\end{lstlisting}

\paragraph{What is established, and what remains.} The harness recovers planted dominance,
complementarity, and silent-failure structure end-to-end, which validates its mechanics but does not
substitute for clinical numbers. On the real cohort, a first cross-attention run yields the
missing-modality degradation of Table~\ref{tab:degradation}. What that run does not yet include is
the per-example views: the failure taxonomy, complementarity, and loud-vs-silent profile on real
predictions, together with the full unimodal-probe sweep, bootstrap confidence intervals, and the
fairness stratification. These are a matter of exporting per-example predictions and pointing the
harness at them, which is compute-cheap at this cohort size, and a canonical rerun with
patient-level cross-validation in place of the single held-out split is in progress.

\section{Modality Asymmetry Beyond Inference Time}
\label{sec:lessons}
The asymmetry this framework measures at inference time is not confined to inference time, and
noticing that sharpens why the framework is needed. Echocardiography is the harder modality to work
with at every stage, not only the harder one to acquire in a clinic. No off-the-shelf echo-to-ECG
temporal join exists, so windowing, tie-breaking, and leakage-safe splits are all bespoke, and DICOM
volume makes echo the heavier modality in development as well. The ECG, cheap and ubiquitous at the
bedside, is correspondingly easy on both axes.

The two asymmetries therefore point the same way: the modality that is expensive to obtain at
inference is also the one whose infrastructure is expensive to build, which is precisely why a system
is liable to be deployed without it and why knowing how it fails without it matters. We read this
correspondence as a deployment finding in its own right, and as a structural feature of the path from
a published cardiac foundation model to a screening tool rather than an accident of this project.

\section{Limitations}
\label{sec:limitations}
Several constraints bound our conclusions. MIMIC-IV is a single US academic center, so degradation
and fairness patterns may not transfer across institutions or acquisition hardware; that is an
external-validation gap we cannot close here. By design we never fine-tune the backbones, so
absolute accuracy is a lower bound on what the representations could achieve, and our claims
concern \emph{relative} degradation under missing modalities rather than state-of-the-art LVEF
estimation. We target structured LVEF and the \mbox{EF\,$\leq$\,40\%} gate only, and
\mbox{EF\,$\leq$\,40\%} prevalence in a paired echo and ECG cohort may be modest, widening AUROC
confidence intervals.

Two further caveats concern the measures themselves. The loud/silent split uses distance from the
\mbox{EF\,$\leq$\,40\%} threshold as a proxy for monitorability, not a probabilistic confidence; a
calibrated head would sharpen it. And with $N{=}2$ the leave-one-out winner equals the
surviving-modality solo, so per-example attribution is partly degenerate and is most informative
for $N{\ge}3$.

A demographic audit inherits MIMIC's known gender- and race-curation limitations, small
per-stratum samples (giving wide or undefined AUROC), and post-hoc confounding by differing
disease prevalence and presentation across groups; observed gaps may reflect the data rather than
the model alone. Finally, the current extraction pipeline embeds one DICOM clip per echo study
instead of pooling across a study's full clip set. This may affect absolute echo-branch accuracy.
The fix is known (per-clip caching with pooling at the data-loader level) and is not yet applied to
the reported numbers.

\section{Discussion}
\label{sec:discussion}
The deployment question reframes multimodal value. What matters is less how much fusion adds when
both modalities are present than which modality is responsible, and whether the model fails loudly
or silently when one is gone. Because the framework assumes only a mask-aware predictor, it
transfers to the next echo or ECG backbone without change; the evaluation outlives the model it was
built to evaluate, which is why we release an instrument rather than a checkpoint. Our validation
study shows why the two axes have to be measured rather than assumed: echo-dominance was robust
across seeds, yet the loud-vs-silent rates of the two modalities were comparable, so value and
detectability need not coincide. Design philosophies for epistemic virtue in clinical
AI, such as BODHI~\citep{bodhi2026}, therefore need this kind of measurement: calibrated humility
presupposes knowing which failures are visible in the first place. Future work extends the
instrument along that same line: further modalities, calibrated per-example profiles, and the same
measurement repeated across datasets and hospital settings, where a failure profile that shifts
between institutions is what tells a deployed system when to defer rather than answer. Designing for
humility needs measurement of that kind, not assertion of it.


{\small
\bibliographystyle{plainnat}
\bibliography{references}
}

\end{document}